\documentclass[sigconf]{acmart}
\usepackage{booktabs} 

\usepackage{mathtools}
\usepackage{graphicx}
\usepackage{amsmath,amssymb} 
\usepackage{color}
\usepackage{subcaption}
\usepackage{url}
\usepackage{balance} 

\begin{document}
\setcopyright{acmcopyright}
\acmDOI{10.1145/3240508.3240570}
\acmISBN{978-1-4503-5665-7/18/10}
\acmConference[MM '18]{2018 ACM Multimedia Conference}{October 22--26, 2018}{Seoul, Republic of Korea}
\acmYear{2018} 
\copyrightyear{2018} 
\acmPrice{15.00}
\acmBooktitle{2018 ACM Multimedia Conference (MM '18), October 22--26, 2018, Seoul, Republic of Korea}

\fancyhead{}
\title{Real-time 3D Face-Eye Performance Capture of\\ a Person Wearing VR Headset }

\author{Guoxian Song}
\affiliation{%
	\institution{Nanyang Technological University}
}
\email{guoxian001@e.ntu.edu.sg}

\author{Jianfei Cai}
\affiliation{%
	\institution{Nanyang Technological University}
}
\email{asjfcai@ntu.edu.sg}

\author{Tat-Jen Cham}
\affiliation{%
	\institution{Nanyang Technological University}
}
\email{astjcham@ntu.edu.sg}

\author{Jianmin Zheng}
\affiliation{%
	\institution{Nanyang Technological University}
}
\email{ASJMZheng@ntu.edu.sg}

\author{Juyong Zhang}
\authornote{The Corresponding author.}
\affiliation{%
	\institution{University of Science and Technology of China}
}
\email{juyong@ustc.edu.cn}

\author{Henry Fuchs}
\affiliation{%
	\institution{University of North Carolina at Chapel Hill}
}
\email{fuchs@cs.unc.edu}

\renewcommand{\shortauthors}{G. Song et al.}

\begin{abstract}
Teleconference or telepresence based on virtual reality (VR) head-mount display (HMD) device is a very interesting and promising application since HMD can provide immersive feelings for users. However, in order to facilitate face-to-face communications for HMD users, real-time 3D facial performance capture of a person wearing HMD is needed, which is a very challenging task due to the large occlusion caused by HMD. The existing limited solutions are very complex either in setting or in approach as well as lacking the performance capture of 3D eye gaze movement. In this paper, we propose a convolutional neural network (CNN) based solution for real-time 3D face-eye performance capture of HMD users without complex modification to devices. To address the issue of lacking training data, we generate massive pairs of HMD face-label dataset by data synthesis as well as collecting VR-IR eye dataset from multiple subjects. Then, we train a dense-fitting network for facial region and an eye gaze network to regress 3D eye model parameters. Extensive experimental results demonstrate that our system can efficiently and effectively produce in real time a vivid personalized 3D avatar with the correct identity, pose, expression and eye motion corresponding to the HMD user.
\end{abstract}



\maketitle

\section{Introduction}
This paper considers the problem for only using a commodity RGB camera to do real-time 3D facial performance capture of a person wearing a virtual reality (VR) head-mount display (HMD). Although 3D facial performance capture alone is a well studied problem ~\cite{thies2016face,li2015facial,Garrido:2016,Cao:2015}, which aims to track and reconstruct 3D faces from 2D images, performance capture of faces with HMDs is a new problem and it only emerged recently due to the increasing popularity of HMDs such as Oculus~\cite{Oculus} and HTC Vive~\cite{HTCVive} and rapid development of various VR applications. One particular application we consider here is teleconference or telepresence, where two or multiple users wearing VR HMDs can have immersive face-to-face communications in a virtual environment. 

\begin{figure}[t]
	\includegraphics[width=1.0\linewidth]{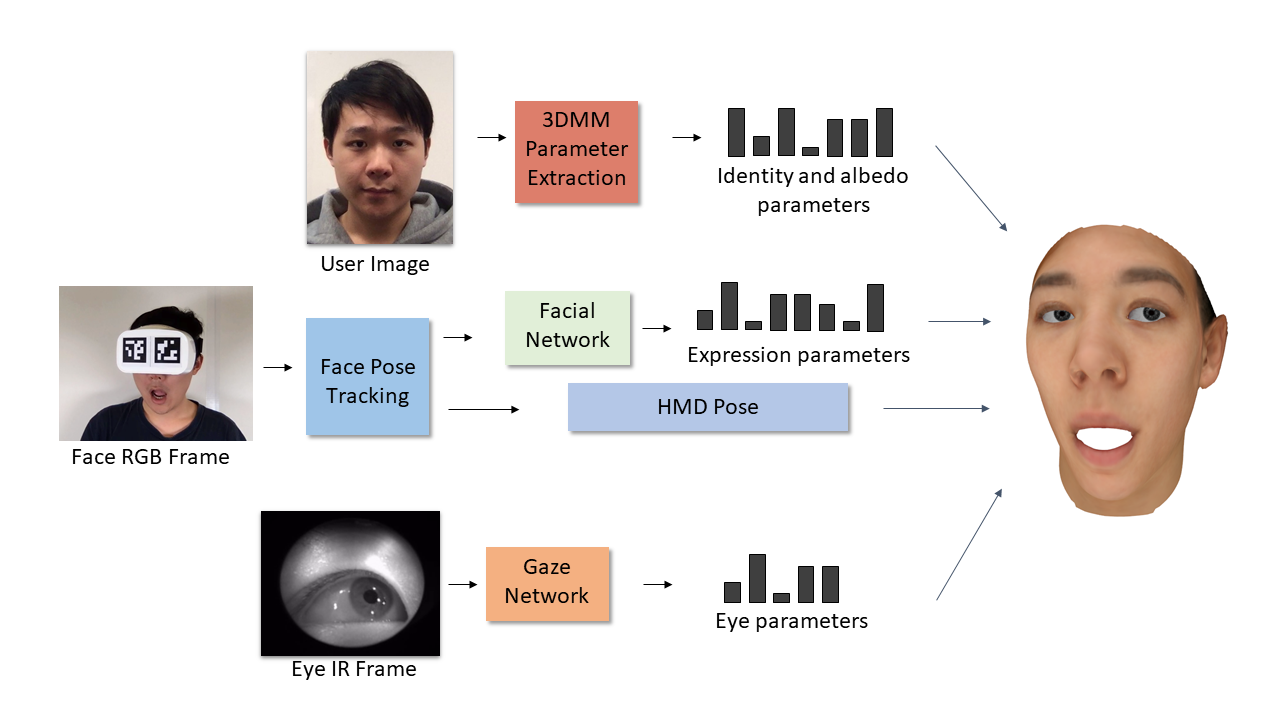}
	\caption{An overview of our system. 3DMM Parameter Extraction module: to extract the personal 3DMM identity and albedo information given one frontal face image, which only needs to be done once and offline. Face Pose Tracking module:  track face pose and locate face region based on ArUco Marker detection~\cite{Garrido}. Facial Network module: given the cropped face region with HMD, to regress the 3DMM expression parameters. Eye Network module: given an eye IR image, to regress the 3D eye model parameters. }
	\label{fig:overview}
\end{figure}

However, such a task is extremely challenging, since more than 60 percent face including all upper face landmarks are occluded by HMD (see the input image in Fig.~\ref{fig:overview}). The largely occluded faces in HMD scenarios make the face performance capture problem intractable for almost all the existing landmark based methods~\cite{thies2016face,Guo20173DFaceNet,li2015facial}, which explicitly leverage on the tracked landmarks to locate the face and recover the full facial motion,

Fortunately, there have been are several advances on this challenging problem. For example, Li et al.~\cite{li2015facial} integrated RGB-D and strain sensors into VR HMD for real-time facial expressions animation, but it requires special hardware setup and tedious calibration for each step, and it cannot automatically digitize a personalized 3D avatar. Thies~\cite{DBLP:journals/corr/ThiesZSTN16} et al. proposed 2D facial reenactment framework for HMD users using RGB-D camera and carefully optimized parallel computing, which is fairly complex. Unfortunately, none of those efforts included the performance capture of 3D eye gaze movement, which limits their performance in producing compelling digital presence. Note that \cite{DBLP:journals/corr/ThiesZSTN16} only classifies eye gaze information to retrieve the corresponding 2D eye image regions in database for producing non-occluded face image by image composition, and it cannot produce a personalized avatar with 3D eye gaze movement. In summary, the existing limited solutions are very complex either in setting or in approach and lack the performance capture of 3D eye gaze movement.

This motivates us to look for a better solution that should be simple in setting, less complex in approach and take into account 3D eye gaze movement. Recently, Guo et al. \cite{Guo20173DFaceNet} proposed a convolutional neural network (CNN) based learning framework for real-time dense facial performance capture, based on the well-known 3D morphable model (3DMM) representation, which encodes a 3D face into three sets of parameters: facial identity, expression and albedo. Leveraged on the large-scale training data, \cite{Guo20173DFaceNet} can train a powerful network to regress the 3DMM parameters as well as pose and lighting parameters in real time given RGB inputs. 

This triggers us to propose a CNN based solution with 3DMM representation for 3D face-eye performance capture of HMD users. Particularly, we develop a facial network to predict facial parameters and an eye gaze network to predict the eye motion information. However, there are three challenges we face here. First, there is no public dataset available that contains HMD facial images and their corresponding 3DMM parameter labels. 
Note that \cite{olszewski2016high} also uses CNN to regress facial parameters, but it needs professional artists to assist on generating labelled training data, which is very costly and time-consuming. Second, given limited observation on face from the input RGB image, it is hard for a network to learn and predict all the 3DMM and pose parameters. Third, there is no training data available for learning a 3D eye gaze network. For the first challenge, we propose to use a data synthesis approach to solve it. Particularly, we synthesize images of face wearing HMD from non-occluded face images. For the second challenge, we adopt a divide-and-conquer strategy. Particularly, for an HMD user, we can obtain its 3DMM identity and albeo parameters in advance from a single full face RGB image, which only needs to be done offline once. During online operation, we
use ArUco Marker~\cite{Garrido} to track face pose and the facial network only needs to regress the expression parameters from the input HMD image. For the third challenge, we make use of the existing eye tracking techniques, which have been integrated into many commercial HMD products such as \cite{Fove,SMI,TobiiPro}. The existing eye tracking-HMDs can provide the captured IR eye images and the corresponding eye gaze information, which will be used to train our eye gaze network to regress the 3D eye model parameters. Fig.~\ref{fig:overview} gives an overview of the proposed system.

The main contributions of this paper are threefold.
\begin{itemize}
\item We construct an HMD face dataset that contains synthetic HMD-occluded facial images with 3DMM expression labels (extracted from full face observations).
    
\item We construct a VR IR eye dataset with various gaze angles. We intend to release both datasets for research purposes. To the best of our knowledge, there is no such dataset currently available in public.
    
\item We develop a face-eye performance capture system that integrates the 3D parametric face-eye models. Extensive experimental results demonstrate that our system can efficiently and effectively produce in real time a vivid personalized 3D avatar with the correct identity, pose, expression and eye motion corresponding to the HMD user.
\end{itemize}

\section{Related Work}
In this section we review the existing methods on 3D face and eye reconstruction and the facial performance capturing methods on HMD faces, which are closely related to this research, as well as the developments of HMD devices.

\textbf{3D face and eye reconstruction.} In the context of 3D facial reconstruction, early works focus on the optimization to minimize the difference between the input RGB images and the reconstructed model appearances. 3D morphable model (3DMM) is the most well-known parametric 3D face model, which disentangles a 3D face into three sets of parameters: facial identity, expression and albedo. Oswald et al.~\cite{aldrian2013inverse} proposed a complete framework for face inverse rendering with 3DMM. Recently, with the development of deep learning technology, Guo et al.~\cite{Guo20173DFaceNet} presented a coarse-to-fine CNN framework for real-time dense textured 3D face reconstruction based on 3DMM with RGB inputs. For unconstrained images (often with occlusions), Saito et al.~\cite{DBLP:journals/corr/SaitoLL16} regressed the 3D face parameters from the non-occluded facial part via semantic segmentation. Yu et al.~\cite{DBLP:journals/corr/abs-1709-00536} presented a neural network for dense facial correspondences in highly unconstrained RGB images. There are also some 3D face reconstruction works based on RGB-D inputs with depth sensors like Kinect and Prime Sense, which provide aligned depth and colour streams. For example, Tan et al.~\cite{Tan:2017} introduced a cost-efficient framework to align point clouds from two depth cameras for high quality 3D facial reconstruction in real time. All these 3D face reconstruction methods cannot handle the HMD face images where the face is largely occluded by HMD.

For eye reconstruction, the eye ball is a complex organ comprised of multiple layers of tissues with different transparency and reflectance properties. Wood et at.~\cite{Wood:2016:LAG:2857491.2857492} presented UnityEyes for rapid synthetic eye data  generation based on high resolution 3D face scan and complex eyeball materials. Recently, Wood et at.~\cite{Wood:2016:MME:3059079.3059098} proposed a 3D morphable model for eye region reconstruction from a RGB image.

\begin{figure}
	\includegraphics[width=.9\linewidth]{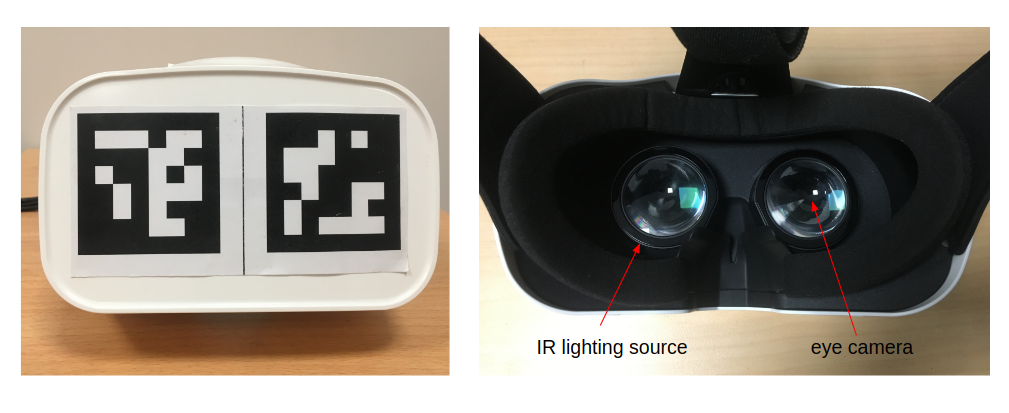}
	\caption{VR HMD device with ArUco markers and eye IR camera.}
	\label{fig:devices}
\end{figure}

\textbf{HMD devices.}
For over two decades, wearable HMDs are widely used in various entertainment and education applications. Foreseeing the tremendous opportunities, eye tracking techniques have recently been integrated into commercial HMD products such as \cite{Fove,SMI,TobiiPro}. For example, a Japanese startup company Fove~\cite{Fove} embedded infrared cameras to track eye movements and released the first commercial eye-tracking VR HMD. Soon after that, Tobii, the largest eye-tracking company collaborated with the well-know VR company HTC Vive~\cite{HTCVive} and developed eye-tracking VR HMDs Tobii Pro~\cite{TobiiPro}.
Meanwhile, on the research side, there is a strong interest in reconstructing the VR rendering environment involved with the users wearing HMD.  Very recently, a team of researchers at Google company started a headset removal project, which aims to remove the large occlusion regions caused by the HMD devices, for virtual and mixed reality based on HTC Vive incorporated with SMI~\cite{SMI} eye-tracking technology~\cite{googleHMD}. Besides, Pupil Labs~\cite{PupilLabs} has developed open source software and accessible eye tracking hardware for existing HMDs, like Oculus~\cite{Oculus} and Hololens~\cite{Hololens}, etc. All together, there efforts make eye tracking HMDs production and distribution much easier and more prevalent than ever. Our system uses Fove 0 HMD (see Fig.~\ref{fig:devices}) to map video recordings of face and eyes to a personalized 3D face-eye model.

\textbf{Facial performance capture of faces with HMD.} There are only a few recent works trying to attack the problem of facial performance capture of HMD faces. Li et al.~\cite{li2015facial} proposed the first VR HMD system for fully immersive face-to-face telepresence. They attached an external depth sensor to HMD to capture the visible facial geometry while wearing HMD, and they estimated facial expression of the occluded regions by measuring the signals of electronic strain sensors attached to the inside of HMD. The resultant expression coefficients are then used to animate virtual avatars. Later on, Olszewski et al.~\cite{olszewski2016high} proposed an approach for an HMD user to real-time control a digital avatar, which is manually created by professional artists. Similarly, they mounted a custom RGB camera for mouth tracking. In addition, a convolutional neural network is used to regress from mouth and eye images to the parameters that control a digital avatar. While the system can provide high-fidelity facial and speech animation, the system is limited to specifically created avatars by professional artists and does not consider the eye gaze movement which is important for vivid presences. Thies~\cite{DBLP:journals/corr/ThiesZSTN16} proposed a 2D facial reenactment framework for HMD users with RGB-D camera. Their method is an optimization based solution, carefully designed for parallel computing in order to achieve real-time performance, which is highly computational complex.

The differences between our approach and the existing methods mainly lie in three aspects: setting, approach, and eye motion capture. In terms of setting, \cite{olszewski2016high} requires the most complex setting, including mounting strain sensor and RGB-D camera on HMD and the tedious calibration process for each subject. \cite{olszewski2016high} also requires mounting RGB camera on HMD. \cite{DBLP:journals/corr/ThiesZSTN16} needs an external RGB-D camera to capture the face of an HMD user. In contrast, our system only needs an external RGB webcam without any calibration, which is the simplest one in setting. In terms of approach, both \cite{olszewski2016high} and \cite{DBLP:journals/corr/ThiesZSTN16} are optimization based solutions, which require intensive computation, while our solution is based on CNN and it only needs one forward pass during online operation, which is much faster. Although \cite{olszewski2016high} also uses CNN to regress facial parameters, they need professional artists to assist on generating labelled training data. On the contrary, we use data synthesis for training data generation, which does not need any human intervention. In terms of eye motion capture, none of the existing works capture 3D eye gaze movement. Although \cite{DBLP:journals/corr/ThiesZSTN16} takes into the eye gaze information, the classified eye gaze information is only used to retrieve the corresponding 2D eye image regions for the subsequent image composition. In contrast, our system produces a personalized avatar with 3D eye gaze movement.

\section{Overview}
Figure \ref{fig:overview} shows the overall diagram of our proposed 3D facial performance capturing system for a person wearing HMD. The entire system consists of three paths located from top to bottom, respectively, as shown in Figure \ref{fig:overview}. The top path is to extract the personal 3DMM identity and albedo information given one frontal face image, using the inverse rendering optimization approach~\cite{Guo20173DFaceNet}, which only needs to be done once and offline. During the online operation, at each time instance, the middle path takes in a face RGB frame and regresses the corresponding 3DMM face expression parameters, and the bottom path takes in an eye IR image frame and regresses the corresponding eye model parameters. The final output of our system is a personalized 3D avatar with the corresponding face pose, face expression and eye gaze. Note that the face pose is directly obtained by the Face Pose Tracking module based on ArUco Marker detection~\cite{Garrido}, which is also used to locate and crop face region so as to align the input face images into the facial network. Overall, the key components of our system are the facial network and the eye network, which respectively regress 3DMM facial expression parameters and 3D eye model parameters in real time. In the next sections, we will describe how to train these two networks in detail.

Particularly, our system is based on the FOVE 0 VR HMD, with integrated eye tracking cameras and custom ArUco markers for face tracking. Each of the infrared (IR) cameras inside HMD is with six 940nm wavelength IR LEDs, by which user's eyes can be observed clearly without the need of ambient illumination. It can record 320x240 images of user's eyes at 60 fps. The user's face is recorded by a commodity RGB WebCam at 30fps with a resolution of 1080x720. Our system is much more convenient than that of Olszewski's~\cite{olszewski2016high}, which needs external camera and lighting source equipment attached to the HMD.

\section{Facial Regression Network}
To train the facial expression regression network, the main challenge is that there is no public HMD facial dataset available and it is extremely difficult to obtain the labels of the ground truth expression parameters with highly occluded faces. \cite{olszewski2016high} requires professional animators to assist on producing expression labels, which is costly and time-consuming. In contrast, we use data synthesis to produce pairs of HMD faces and labels, which does not require any human annotation. In the following, we first introduce the 3D parametric face model we adopt, and then describe our data synthesis and network structure in detail.

\subsection{Parametric 3D Face Model}
We use 3D Morphable Model
(3DMM) \cite{blanz19993DMM} as the parametric face model to encode 3D face geometry and albedo.
Specifically, 3DMM describes 3D face shape geometry $S$ and color albedo $C$ with PCA (principal component analysis) as
\begin{align}
S &= \bar{S}+A^{id}x ^{id} + A^{exp}x ^{exp} \\
C &= \bar{C}+A^{alb}x^{alb}
\label{equ:eq_3dmm}
\end{align}
where $\bar{S}$ and $\bar{C}$ denote respectively the 3D shape and albedo of the average face, $A^{id}$ and $A^{alb}$ are the principal axes extracted from a set of textured 3D meshes with a neutral expression, $A^{exp}$ represents the principal axes trained on the offsets between the expression meshes and the neutral meshes of individual persons, and $x^{id}$, $x^{exp}$ and $x^{alb}$ are the corresponding coefficient vectors that characterize a specific 3D face model. In this research, $x^{id}$ and $x^{alb}$ are of 100 dimensions, with the bases $A^{id}$ and $A^{alb}$ from the Basel Face Model (BFM)~\cite{paysan20093d}, while $x^{exp}$ is of 79 dimensions with the bases $A^{exp}$ from FaceWarehouse~\cite{cao2014facewarehouse}.

\subsection{Generating Training Data for Facial Network} \label{sec:face_training}

\begin{figure}
	\includegraphics[width=.9\linewidth]{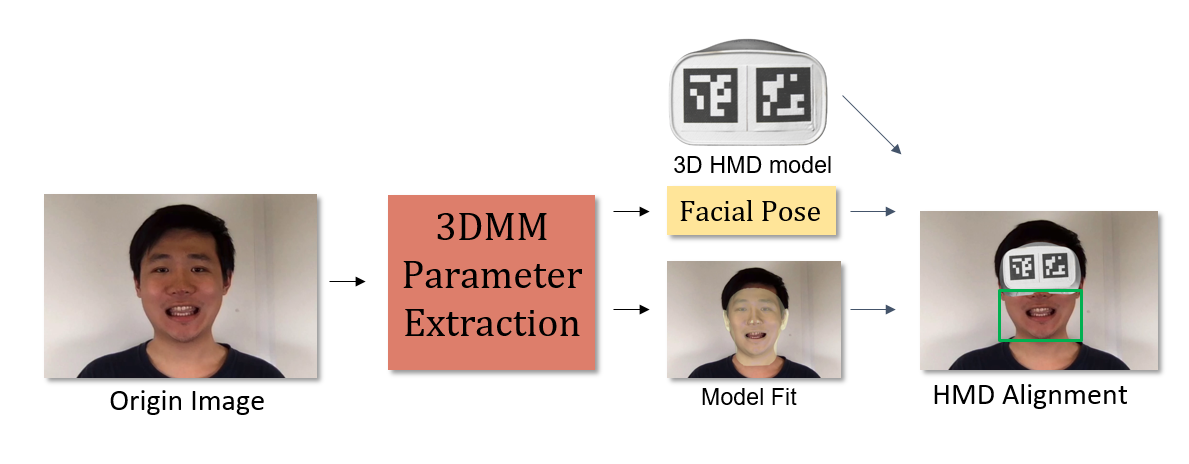}
	\caption{The pipeline for generating facial network training data.}
	\label{fig:HMD_fit}
\end{figure}

As aforementioned, to train the facial network, we need to have the input-output pairs with the input being HMD faces and the output being the corresponding 3DMM facial expression parameters. Our basic idea is to make use of non-occluded faces, for which it is easy to obtain the corresponding expression parameters (treated as ground-truth label), and then add HMD into non-occluded faces by data synthesis to create the corresponding input images. Fig.~\ref{fig:HMD_fit} shows the pipeline for generating facial network training data.

In particular, we record videos of six people (5 male and 1 female) without wearing HMD. To capture subtle variations of a user's expression, each subject firstly performs different expression movements, including opening and closing mouth, smiling and pulling, and then slowly pronounces vowels and a list of 10 sentences from Harvard sentences~\cite{Harvard}. For different pose, subjects are allowed to move their head while performing expression.

For each video frame, we first use the inverse rendering method \cite{Guo20173DFaceNet} to extract the 3DMM coefficients $\{x^{id},~x^{alb},~x^{exp}\}$ and the pose parameters $\{R,t,s\}$, where $R,t,s$ denotes rotation, translation and scale respectively. With the pose parameters, a 3D face mesh $S$ specified by $\{x^{id},~x^{alb},~x^{exp}\}$ can be projected into the image plane according to the weak perspective projection model (see `model fit' in Fig.~\ref{fig:HMD_fit}):
\begin{align}
p_{i}
=s
\left(
\begin{array}{ccc}
1 & 0 & 0\\
0 & 1 & 0\\
\end{array}
\right)
R v_{i}+t, \ v_{i}\in S
\label{equ:projection}
\end{align}
where $p_{i}$ and $v_{i}$ are the projected 2D point (pixel) in the image plane and the corresponding 3D point on the face mesh $S$.

To synthesize photo-realistic HMD facial images, we use the 3D scanner Artec Space Spider \cite{Artec} with 0.05mm accuracy to scan the HMD and get the 3D HMD model and texture. Then, for each frame, we project the scanned 3D HMD into the image according to its face pose (see `HMD alignment' in Fig.~\ref{fig:HMD_fit}). With the synthesized HMD face, finally we apply ArUco Marker detection~\cite{Garrido} (same as the `Face Pose Tracking' module during the online operation in Fig.~\ref{fig:overview} for training-testing consistency) to locate the visible face part (the green bounding box in Fig.~\ref{fig:HMD_fit}), which is used as the input to the facial network to regress the expression parameters. Note that, to eliminate the difference caused by illumination, we mask the HMD part $M_{HMD}$ as black pixels in both training and testing images:
\begin{align}
M_{HMD} = \{ \Pi (Rv+t)| \forall v \in \Omega_{HMD} \}
\label{equ:hmd_mask}
\end{align}
where $\Pi =s
\left(
\begin{array}{ccc}
1 & 0 & 0\\
0 & 1 & 0\\
\end{array}
\right) $ and $v$ denotes a point on the HMD mesh $\Omega_{HMD}$.

\subsection{Facial Network Architecture}
\begin{figure}
	\includegraphics[height=2in]{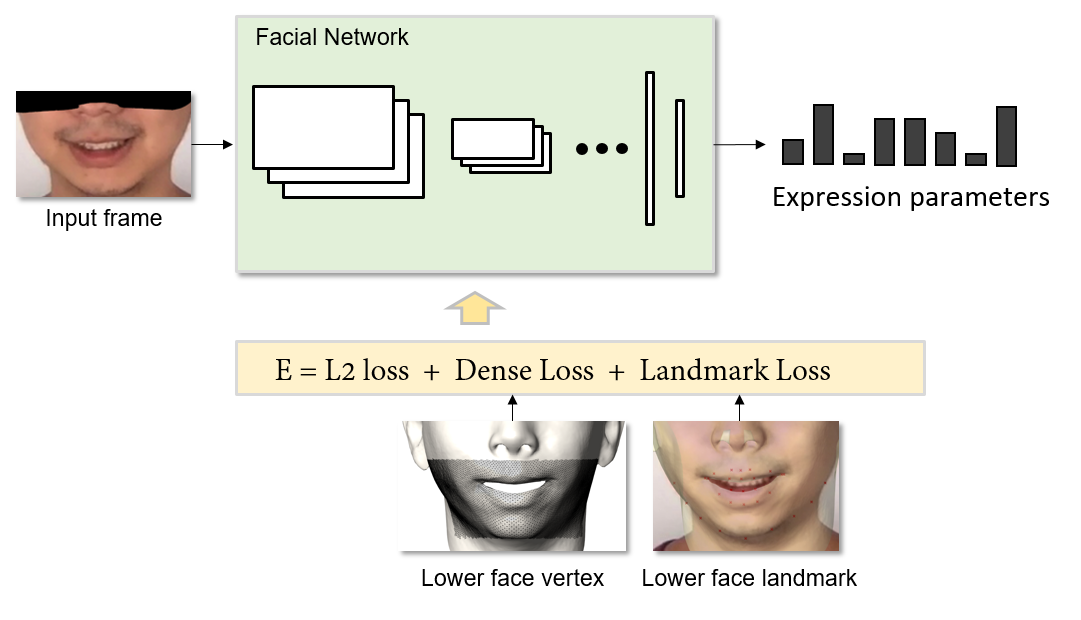}
	\caption{Facial network architecture.}
	\label{fig:facial_network}
\end{figure}

With the obtained cropped face images $\tilde{I}$ and their corresponding labels of the expression parameters $x^{exp}$, we are now ready to train a facial regression network to learn a mapping: $x^{exp}_* = \psi_{F}(\tilde{I})$, where $x^{exp}_*$ is the network predicted expression parameter. Fig.~\ref{fig:facial_network} shows the overall architecture of the facial network.

Specifically, the input cropped frame is of size 112x224. We modify ResNet-18 structure \cite{DBLP:journals/corr/HeZRS15} to fit our frame dimension and also change the last layer to regress the expression parameter of 79 dimension. We define the loss function for the facial network as
\begin{align}
E= \left\lVert \psi_{F}(\tilde{I}) - x^{exp}  \right\rVert_{L_{2}} + \omega_{d}  \sum_{v_i^{*} \in \Omega_{v}^{*}, v_i \in \Omega_{v}} \left\lVert {v_i}^{*} - {v_i} \right\rVert_{L_{2}} + \nonumber \\
\omega_{l} \sum_{v_i^{*} \in \Omega_{l}^{*}, v_i \in \Omega_{l}} \left\lVert \Pi (R{v_i}^*+t) - \Pi (R{v_i}+t) \right\rVert_{L_{2}}
\label{equ:loss}
\end{align}
where $w_d$ and $w_l$ are tradeoff parameters, $\Omega_{v}^{*}$ and $\Omega_{v}$ are the visible vertex set on the reconstruction meshes defined as
\begin{align}\label{equ:mesh_rec}
\Omega_{v}^{*} \subset \bar{S}+A^{id}x^{id} + A^{exp}x ^\psi_{F}(\tilde{I}) \nonumber \\
\Omega_{v} \subset \bar{S}+A^{id}x ^{id} + A^{exp}x^{exp} ,
\end{align}
and $\Omega_{l}^{*}$ and $\Omega_{l}$ are the corresponding 3D landmark sets around the mouth region. Eq.~\eqref{equ:loss} essentially consists of three terms: the first term is a common $L2$ loss on the expression parameter; the second term is a dense loss term to enforce the dense consistency on the visible part of the 3D meshes; and the third term is a landmark loss term to emphasize the consistency on the projected 3D landmark points.

\section{Eye Regression Network} \label{sec:eye}
The purpose of the eye regression network is to regress the 3D eye model parameters, given an IR eye image. In this section, we will first introduce the adopted 3D eye model, and then describe how we prepare the training data and the developed eye regression network architecture.
\begin{figure}[h]
	\includegraphics[width=.9\linewidth]{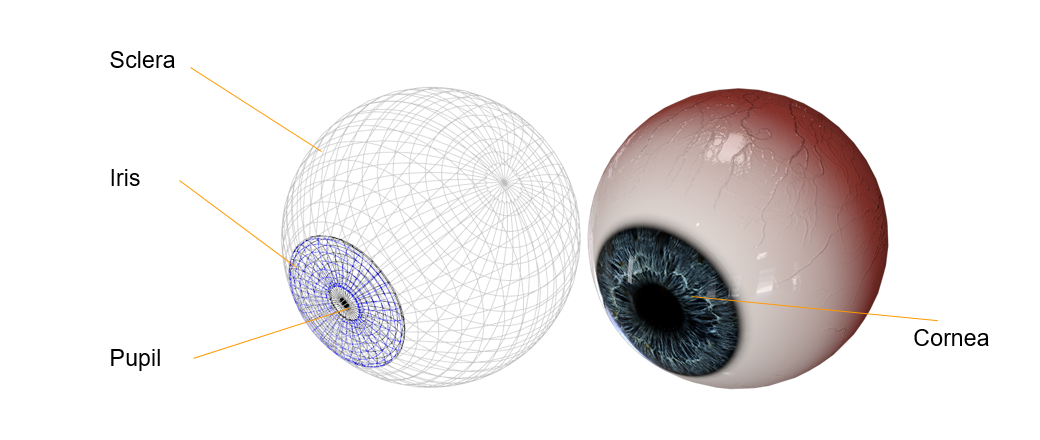}
	\caption{The adopted multi-layer parametric eye model.}
	\label{fig:eye_model}
\end{figure}

\textbf{Parametric Eye Model.} Eye ball is a complex organ comprised of multiple layers of tissues: sclera, cornea, pupil and iris, as shown in Fig.~\ref{fig:eye_model} \cite{DBLP:journals/corr/WoodBZSRB15}.
Cornea is the transparent layer forming the front of the eye with refractive n=1.376. The inner part composed by a pupil (black) and iris (colourful). Sclera is the white outer layer of the eyeball which is connected with the cornea and covered by the veins. We get the basic eye model from the artist store~\cite{EyeData}, which has 20 different iris colours, and parameterize it to make shape variable in pupillary dilation. With the 3D eye model, to render a 3D gaze, we only need to specify the eyeball gaze direction, position, iris colour and pupil size. Considering the iris color can be specified offline in advance for a user, during the online operation, the eye regression network just needs to regress the 5 parameters: pitch, yaw, pupil size, pupil centre $(x,y)$, where the first two parameters are the angles specifying the gaze direction.

\textbf{Generating training data for eye network.}
To train the eye regression network, we need the input-output pairs with the input being eye IR images and the output being the corresponding five eye-model parameters. Since there is no such labeled eye IR image dataset available, we construct the dataset from the scratch. In particular, we record sequences of 7 users' eyes (5 male and 2 female) from the embed IR cameras of HMD while the users are watching a black dot moving in the virtual scene. The black dot moves in a circle and also moves forward and backward in 8 directions to create gaze variations and pupil size variations.

For obtaining the gaze orientation label, we directly use the gaze vector provided by the FOVE HMD device. Particularly, we first calibrate eye gaze for each subject using the default calibration program from the device, and then record each IR frame with the gaze vector returned from the device.

\begin{figure}[h]
	\includegraphics[height=1in]{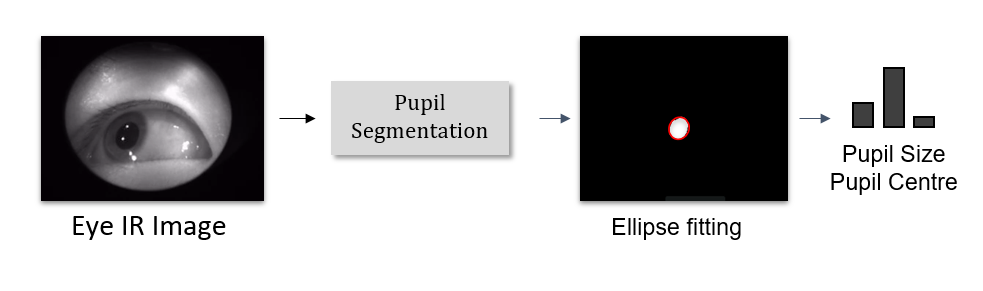}
	\caption{Image processing pipeline for pupil segmentation.}
	\label{fig:ellipse_fit}
\end{figure}

For obtaining the labels of pupil size and position, we use the conventional image processing techniques to segment pupil. Fig.~\ref{fig:ellipse_fit} shows the pipeline for pupil segmentation. Specifically, given an eye Image, we use the method~\cite{pupil_tracking} to calculate an accumulated value of a fixed-size area 5x5 for each pixel to find the darkest point, which is deemed to be in the pupil region. Then, we set the darkest point as a seed, and use the classical watershed segmentation method to segment the pupil. Considering there are 6 IR lighting points that often cause holes or discontinuity for segmentation, we further refine the result using hole filling and morphology techniques~\cite{segmentation}. After that, we use the conventional ellipse fitting method~\cite{ellipse} to find an optimal ellipse to match the segmented contour. Finally, we label the ellipse centre as the pupil centre, and use the major axis length as the pupil size.

\begin{figure}[h]
	\includegraphics[height=1in]{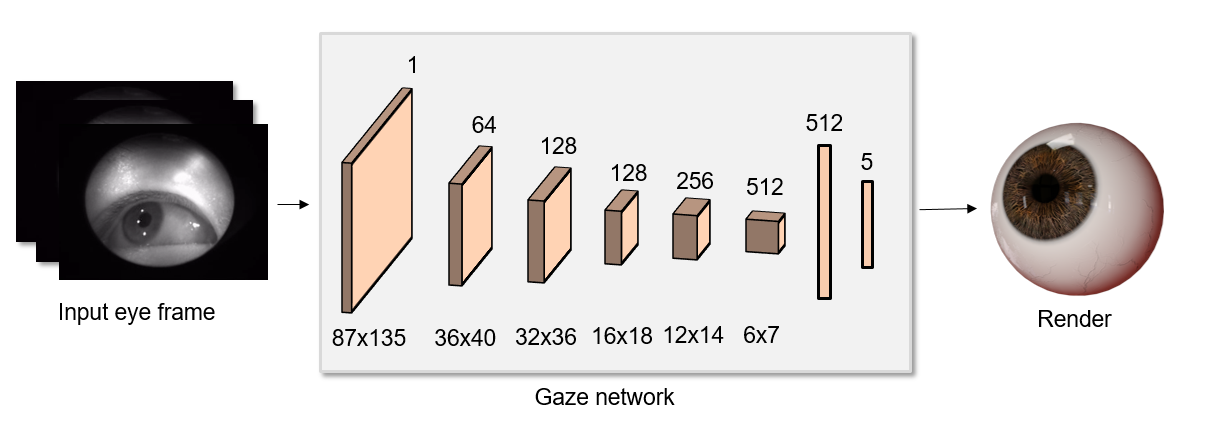}
	\caption{Architecture for the developed eye gaze network.}
	\label{fig:architecture_gaze}
\end{figure}

\textbf{Eye Network Architecture.}
Fig.~\ref{fig:architecture_gaze} shows the architecture of the developed eye gaze network. In particular, an eye IR image is first rescaled to 87x135 before input into the eye network,  and then we apply a chain of CNN operations: 11x11 and 5x5 convolutional layer with reLU, 2x2 max pooling, followed by 5x5 convolutional layer and 2x2 max pooling, and a final fully connected layer with 5 parameters output. We use L2 loss to regress the five parameters: pitch, yaw, pupil size, and pupil centre $(x,y)$.

\section{Experimental Results}

\subsection{Implementation}
For the facial network, we choose 8608 recorded images with various facial expressions. For each image, as described in Section~\ref{sec:face_training}, we generate an HMD mask on the image and crop the lower face region of size 120x230 according to the face pose. For data augmentation, we further randomly crop 10 times for each facial 120x230 image into 112x224. In this way, we generate 86080 112x224 images for training. The learning rate is initialized as 0.001 and decays with a base of 0.9 at every epoch for total 70 epochs. The tradeoff parameters $ w_l$ and$ w_d$ are set as 1 and $1 \cdot 10^{-6}$.

For the eye gaze network, we collect 18806 frames for 7 subjects from IR cameras and record the gaze labels.
The pupil size and centre labels are generated using the image processing pipeline described in Section~\ref{sec:eye}. We calibrate each eye IR camera and undistort the image. For data augmentation, we mirror left eye image to right image and get in total 14 groups. And we further randomly crop 10 times of each eye image into a size of 87x135. In this way, we obtain totally 376120 images for the eye gaze network. The learning rate is initialized as 0.0001 and decays with a base of 0.96 at every half epoch for total 70 epochs.

The facial and eye networks are trained with a batch size of 64 and 256, respectively, on a PC with 2 GPUs GTX1080. The system is tested with a variety of subjects under different circumstances. The system is very convenient to use for different people since it does not require any user-dependent calibration.

\subsection{Results on Facial Expression Capture}
\begin{figure*}
	\includegraphics[width=0.8\linewidth]{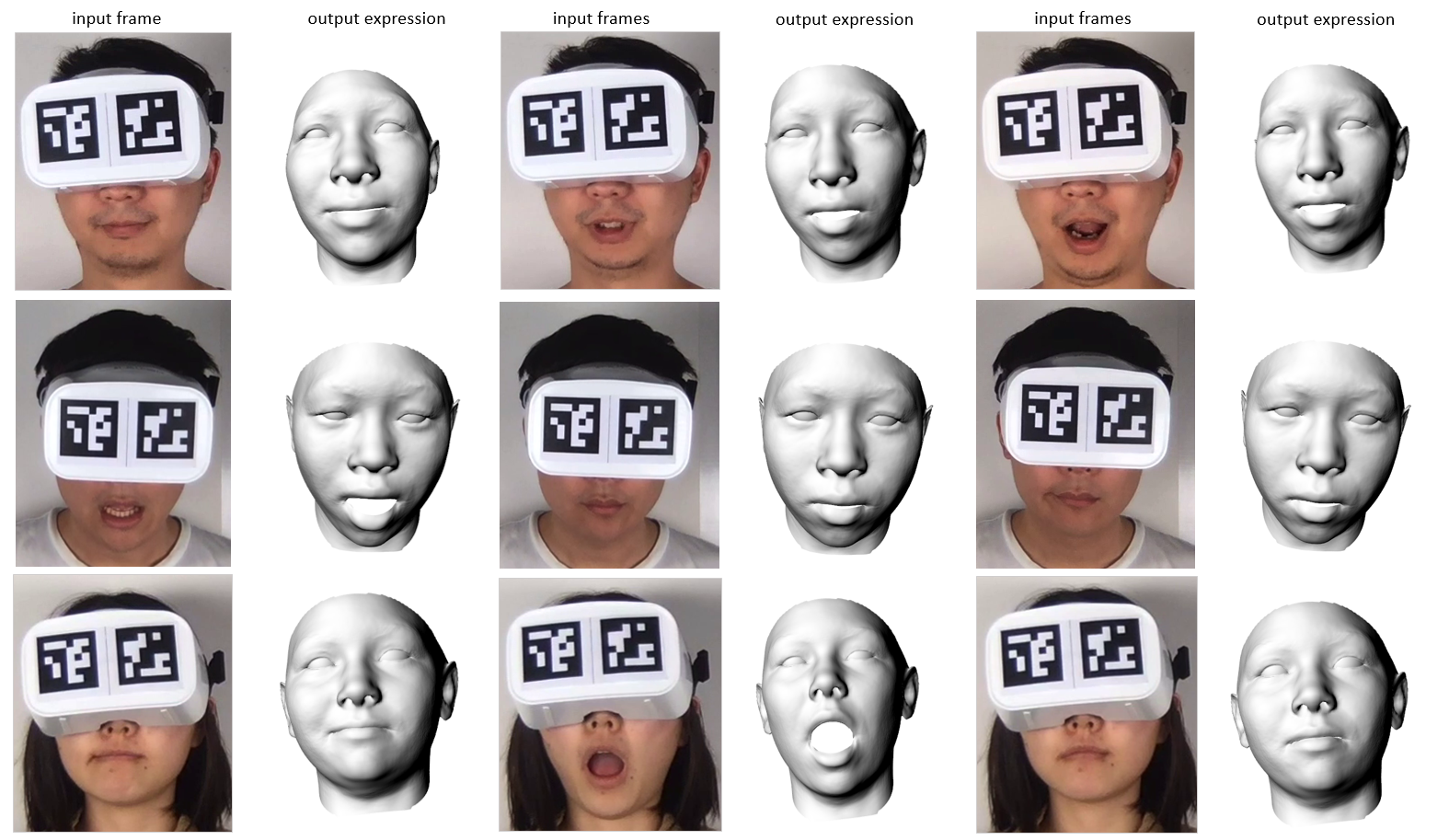}
	\caption{Expression capture results obtained with our facial network on a variety of users. For the real-time facial performance capture, please see our supplementary video.}
	\label{fig:expression}
\end{figure*}

\textbf{Visual results of facial expression capture.}
Fig.~\ref{fig:expression} shows the visual results of facial expression capture using our facial network on a variety of users. We can see that even with faces being largely occluded by HMD, our system can still regress the expression parameter and reconstruct the face shape properly. Our system is quite robust to different people, different expressions and different poses. For the real-time facial performance capture, please see our supplementary video.

\textbf{Quantitative evaluation of facial expression capture.} In order to quantitatively evaluate the proposed HMD facial expression capture, we use synthesized data for testing. Specifically, we simulate 500 test images of one subject, while using the synthesized images of the remaining subjects for training. We use the mean 2D landmark error ($L_2$ distance) as the evaluation metric. For each synthesized HMD face image, since we have the original non-occluded face image, we apply  the facial landmark detector, dlib \cite{dlib09}, on the non-occluded face image, and choose 29 detected landmarks around the lower face region as the `ground-truth' landmarks for evaluation. The predicted 2D landmarks are obtained by projecting the corresponding 3D landmark vertices of the reconstructed 3D face mesh into the image plane. Note that none of the existing facial landmark detection methods can work on HMD face images.

Table~\ref{tab:MLE} lists the mean landmark error results. We compare our method with a simplified version, i.e. only use the $L_2$ loss (the first term in \eqref{equ:loss}) for training. We can see that, compared with the baseline, our method achieves much lower landmark error by introducing the dense loss and the landmark loss in \eqref{equ:loss}. This is because the $L_2$ loss alone cannot distinguish the importance of different expression parameters, and during our experiment, we find out that using combined loss can make network training converge faster.

\begin{table}
	\caption{Quantitative results of facial expression capture.}
	\label{tab:MLE}
	\begin{tabular}{ccl}
		\toprule
		  &Mean landmark error (pixel)\\
		\midrule
		$L_2$ Loss& 3.04\\
		Our method & 0.94\\

		\bottomrule
	\end{tabular}
\end{table}

\subsection{Results on Eye Motion Capture}
\begin{figure*}
	\includegraphics[width=0.8\linewidth]{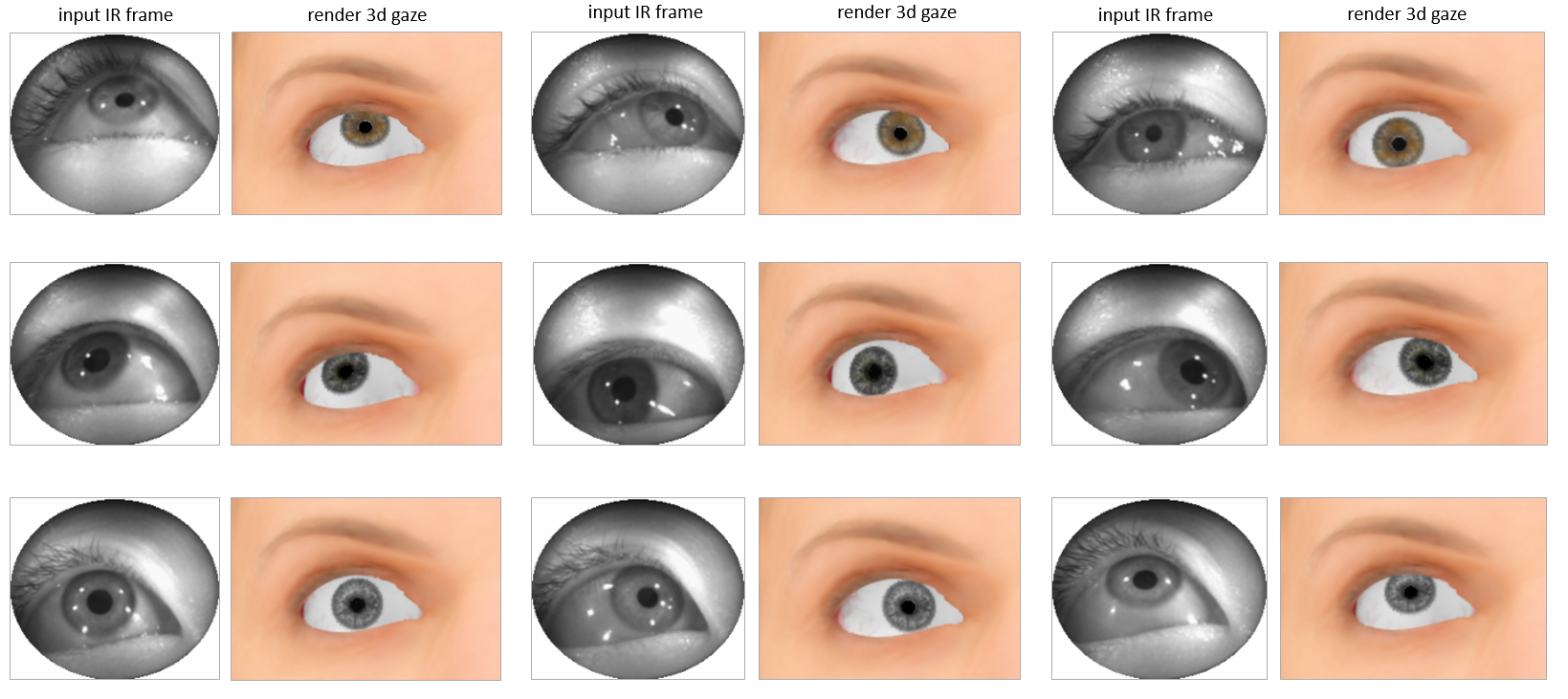}
	\caption{Gaze estimation results obtained by our eye gaze network on different users.}
	\label{fig:gaze}
\end{figure*}

\textbf{Visual result of gaze estimation.} Fig.~\ref{fig:gaze} visualizes our 3D gaze estimation results using our eye gaze network on different users. It can be seen that our system tracks the eye gaze well and is robust to different gaze directions, pupil locations and different users. We also put a sequence for gaze estimation in the supplementary video.

\begin{figure*}
	\includegraphics[width=0.8\linewidth]{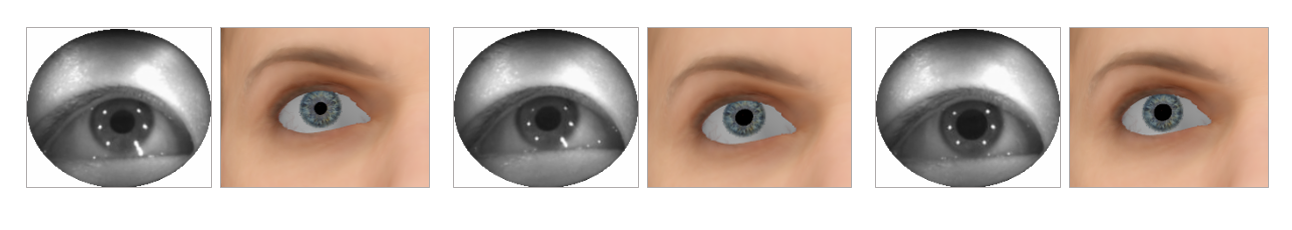}
	\caption{Gaze estimation results on the cases with subtle changes of users pupil size.}
	\label{fig:gaze_dilation}
\end{figure*}

Fig.~\ref{fig:gaze_dilation} shows the results on the cases with subtle changes of users pupil size. As we know, different lighting environment or rending object at different distances for HMD might cause physiological pupillary response that varies the size of the pupil. This pupil's dilation or constriction can be well captured by our eye gaze network via regressing the pupil size, as demonstrated in Fig.~\ref{fig:gaze_dilation}.

\textbf{Quantitative results of eye gaze capture.}
Table~\ref{tab:gaze} shows the quantitative results of eye gaze capture in terms of mean gaze error and runtime. We compare the estimated gaze vectors with the ground truth gaze vectors, which are directly obtained from the HMD device. We choose one subject's left eye as the test sample and keep the rest 13 groups for training. We compare our CNN based solution with an optimization based baseline method, which uses the traditional pupil tracking method~\cite{pupil_tracking} to fit the projected 3D pupil contour to the 2D pupil contour. Particularly, similar to our image processing pipeline shown in Fig.~\ref{fig:ellipse_fit}, the baseline first segments the pupil region, finds the optimal ellipse contour~\cite{ellipse}, and then computes the ratio of the ellipse's major and minor lengths, from which it can obtain the pitch and yaw of the 3D pupil contour~\cite{pupil_tracking}.

\begin{table}
	\caption{Quantitative results of eye gaze capture.}
	\label{tab:gaze}
	\begin{tabular}{ccl}
		\toprule
		 &Mean gaze error &Runtime/frame\\
		\midrule
		Optimization method~\cite{pupil_tracking} &${26.6}^{\tiny o}$ &136.9 ms\\
		Our method &${11.6}^{\tiny o}$ &2.18 ms  \\
		\bottomrule
	\end{tabular}
\end{table}

\begin{figure*}[t]
	\includegraphics[width=0.8\linewidth]{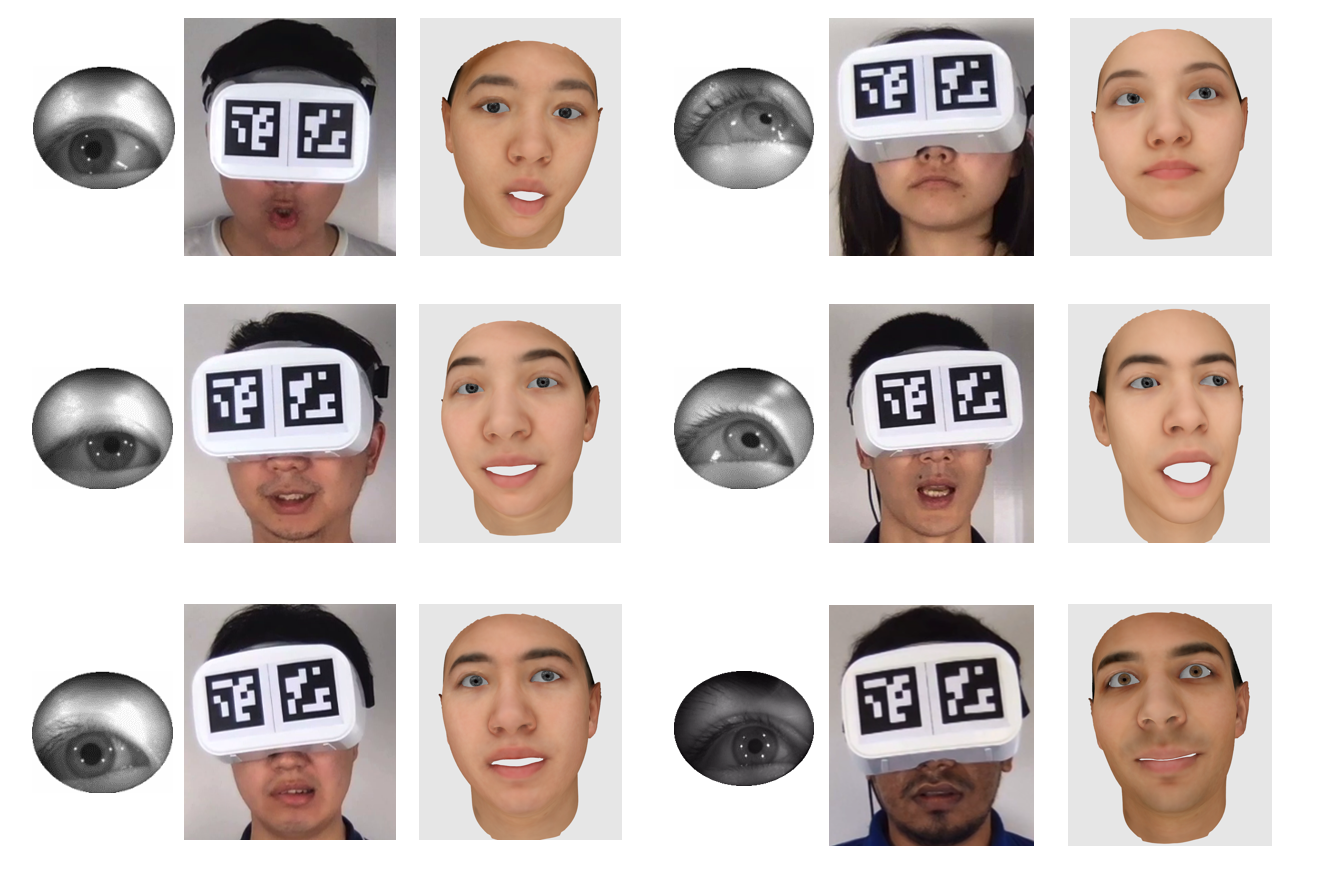}
	\caption{Integrated face-eye performance capture results for various users.}
	\label{fig:combination}
\end{figure*}

From Table~\ref{tab:gaze}, we can see that our method is much faster than the optimization method, since the latter needs to go through a series of image processing while the former only needs to go through one forward pass on the eye gaze network. Moreover, the mean gaze error of our method is much smaller than the baseline.

\subsection{Other Results and Discussions}
\textbf{Integration results.} Fig.~\ref{fig:combination} shows the visual results of the integrated face-eye performance capture on various users with different pose. For front view, please see our video. It can be seen that, given a RGB frame of a user wearing HMD and an eye IR frame, our system can produce in real time a personalized avatar with the corresponding identity, pose, expression and eye motion. The entire system spends 31.5 ms on the facial network and 2.18 ms on the eye gaze network, and overall it can achieve 30 fps.

\begin{figure}[h]
	\includegraphics[width=0.8\linewidth]{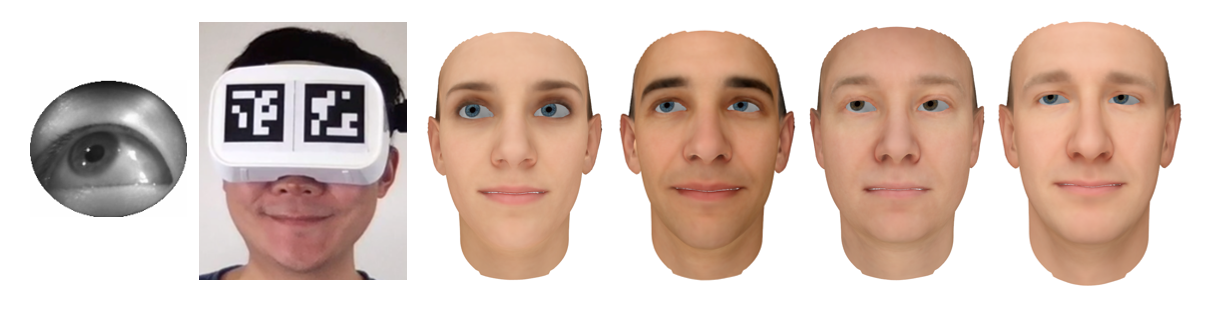}
	\caption{Retarget the face-eye motion to different avatars.}
	\label{fig:retarge}
\end{figure}

\textbf{Face-eye motion retarget.} With the regressed expression and eye parameters, we can easily retarget the face-eye motion to different avatars by using different mesh identity and albedo parameters, as illustrated in Fig.~\ref{fig:retarge}.

\textbf{Limitations.}
There are some limitations in our system. First, for ArUco Marker detection we used to detect the position of HMD, occasionally it might not detect the markers correctly, which will lead to the system failure or flicking. Second, as we only collected data from a limited number of subjects under normal lighting situation, the system may not be robust to users whose appearances are significantly different from the training data. We are planing to collect more data from a wider group of subjects to improve our network robustness. Third, our system does not consider the eyelid movement. It will not be able to regress the gaze when the user is about to close his or her eyes.

We would like to point out that, although the recent studies~\cite{li2015facial,olszewski2016high,DBLP:journals/corr/ThiesZSTN16} attack similar problems, we are not able to give comparisons. This is mainly because of the high complexity of their systems, no available codes, and different system setups. For example, \cite{li2015facial} needs the input from strain sensors. Both \cite{li2015facial} and \cite{DBLP:journals/corr/ThiesZSTN16} requires RGB-D inputs.

\section{Conclusions}
In this paper, we have presented a CNN-based 3D face-eye capture system for HMD users. Our system integrates a 3D parametric gaze model into 3D morphable face model, and it can easily produce a digital personalized avatar from the input of a RGB HMD face image and an IR eye image, with no calibration step. Moreover, to train the facial and eye gaze networks, we collect face and VR IR eye data from multiple subjects, and synthesize pairs of HMD face data with expression labels. Extensive results show that our system is robust in capturing facial performance and 3D eye motion. It provides a promising direction for VR HMD based telepresence.

\section{Acknowledgments}
We thank all reviewers for their valuable comments. We would like to thank Yudong Guo for 3DMM extraction module and also thank Chuanxia Zhen, Deng Teng, Yujun Cai, Eri Ishikawa, Chenqiu Zhao and Ayan Kumar Bhunia for data collection. This research is supported by the BeingTogether Centre, a collaboration between Nanyang Technological University (NTU) Singapore and University of North Carolina (UNC) at Chapel Hill. The BeingTogether Centre is supported by the National Research Foundation, Prime Minister Office, Singapore under its International Research Centres in Singapore Funding Initiative. Also, this research is partially supported by SCALE@NTU.

\bibliographystyle{ACM-Reference-Format}
\balance
\bibliography{sample-bibliography}

\end{document}